\documentclass{article}

\PassOptionsToPackage{numbers, compress}{natbib}

 \usepackage[final]{nips_2018}

\usepackage[utf8]{inputenc} 
\usepackage[T1]{fontenc}    
\usepackage{hyperref}       
\usepackage{url}            
\usepackage{booktabs}       
\usepackage{amsfonts}       
\usepackage{nicefrac}       
\usepackage{microtype}      
\usepackage{color}

\usepackage{amsmath,amssymb,bm}
\usepackage{amsthm}
\usepackage{graphicx}
\usepackage{subfigure}
\usepackage{multirow,multicol,caption}
\usepackage{comment}
\usepackage{makecell}
\usepackage{hhline}
\usepackage{cellspace}

\usepackage{tikz}
\usetikzlibrary{arrows,shapes,positioning}
\usetikzlibrary{calc,decorations.markings}
\usetikzlibrary{decorations.pathreplacing}

\newcommand{\beq}{\begin{equation}}
\newcommand{\eeq}{\end{equation}}

\title{FineText: Text Classification via Attention-based Language Model Fine-tuning}

\author{
  Yunzhe Tao$^1$ \quad\quad\quad Saurabh Gupta$^1$ \quad\quad\quad Satyapriya Krishna$^2$ \\
  {\bf Xiong Zhou$^1$ \quad\quad\quad Orchid Majumder$^1$ \quad\quad\quad Vineet Khare$^3$ \;\;\ \ }\\
  $^1$: Amazon Web Services, $^2$: Amazon Search, $^3$: Amazon Alexa\\
  \texttt{\{yunzhet,gsaur,satyapk,xiongzho,orchid,vkhare\}@amazon.com} \\
}

\begin{document}

\maketitle
\begin{abstract}
Training deep neural networks from scratch on natural language processing (NLP) tasks requires significant amount of manually labeled text corpus and substantial time to converge, which usually cannot be satisfied by the customers. In this paper, we aim to develop an effective transfer learning algorithm by fine-tuning a pre-trained language model. The goal is to provide expressive and convenient-to-use feature extractors for downstream NLP tasks, and achieve improvement in terms of accuracy, data efficiency, and generalization to new domains. Therefore, we propose an attention-based fine-tuning algorithm that automatically selects relevant contextualized features from the pre-trained language model and uses those features on downstream text classification tasks. We test our methods on six widely-used benchmarking datasets, and achieve new state-of-the-art performance on all of them. Moreover, we then introduce an alternative multi-task learning approach, which is an end-to-end algorithm given the pre-trained model. By doing multi-task learning, one can largely reduce the total training time by trading off some classification accuracy.
\end{abstract}

\section{Introduction}\label{sec:intro}
In recent years, deep learning approaches have achieved state of the arts on many NLP tasks, such as machine translation \cite{vaswani2017attention}, text summarization \cite{zhou2017selective}, and sentiment analysis \cite{shen2017disan}. However, they are usually trained from scratch. On one hand, training deep networks based on recurrent neural network (RNN) or convolutional neural network (CNN) requires significant amount of manually labeled text corpus and substantial time to converge, which usually cannot be satisfied by the customers. On the other hand, in comparison to NLP tasks, people can get a high performing model on supervised computer vision (CV) tasks with a very small dataset. So far, the most commonly used pre-training strategy for text is using pre-trained word embeddings, \textit{e.g.}, \cite{mccann2017learned,pennington2014glove}, which helps to boost the performance but cannot alleviate the requirement of having sufficiently large labeled data.

Transfer learning methods have been widely used in CV. Pre-trained models that trained on ImageNet dataset can be applied to fine-tuning image classification models, or many other CV applications, on datasets coming from similar but different distributions. With fine-tuning techniques, the target model can perform well even with a small dataset. Moreover, the model training can be completed fairly quickly as well if the dataset size is reasonable. 

However, for NLP applications, fine-tuning has not seen much success apart from using pre-trained word embeddings until very recently. Some works \cite{devlin2018bert,howard2018universal,peters2018deep,radford2018improving} pre-trained a neural language model on large scale datasets and then applied it to the target tasks. With suitable unsupervised pre-training and fine-tuning algorithms, one can get improvement of the performance on various downstream tasks. Language model (LM) is a broadly useful tool for building NLP systems. For a long time, it has been found useful for improving the quality of generated text from machine translation \cite{schwenk2012large,vaswani2013decoding}, speech recognition \cite{arisoy2012deep,mikolov2010recurrent}, and text summarization \cite{filippova2015sentence,rush2015neural}. More excitingly, it has been recognized recently that Neural Language Model can be used as a powerful feature extractor for texts when it is trained on a large amount of unannotated data, \textit{e.g.}, \cite{peters2017semi,peters2018deep,radford2017learning}.

In this paper, we aim to propose an end-to-end fine-tuning algorithm, given the pre-trained language model, on the text classification tasks. The contributions of our work lie in the following aspects:
\begin{itemize}
\item We compare the performance of different language model pre-trained on different datasets, trying to demonstrate what factor of the pre-trained model is the most important to the downstream tasks performance.
\item We adopt a self-attention mechanism to extract significant contextualized features from the pre-trained LM and use those features as the representations to do classification tasks. The experimental results show that we can achieve new state-of-the-arts on all the six datasets we have tested.
\item We introduce the multi-task learning approach, which simultaneously fine-tunes the classifier and pre-trained LM on the target classification datasets, leading to an end-to-end algorithm given the pre-trained model. By doing multi-task learning, we can largely reduce the total training time.
\end{itemize}

\section{Related Work}\label{sec:work}
There has been a recent resurgence on using language models as the starting point to generate contextualized word or sentence embeddings which can later be used, via fine tuning, to train on the specific end applications, \textit{e.g.}, classification, entailment, and so on.

The ELMo model \cite{peters2018deep} introduces a new type of deep contextualized word representations. The representations are learned functions of the internal states of a deep bidirectional language model, which is pre-trained on a large text corpus, \textit{i.e.}, One Billion Word Benchmark dataset \cite{chelba2013one}. Then for a specific downstream task, ELMo can be easily added to existing models and improve the performance. This kind of method is known as \textit{hypercolumn} in the domain of transfer learning \cite{peters2017semi}.

Furthermore, the ULMFiT (Universal Language Model Fine-tuning) algorithm \cite{howard2018universal} is an effective transfer learning method that can be applied to a variety of classification problems in NLP, which introduces techniques that are key for fine-tuning a language model. ULMFiT consists of three steps, namely general-domain LM pre-training, target task LM fine-tuning, and target task classifier fine-tuning. Different from the ELMo model which incorporates the embeddings to the task-specific neural architectures, ULMFiT employs a simple two-block classifier for all downstream tasks. The input of the classifier is a concatenation of the hidden state at the last time step of the document with both the max-pooled and the mean-pooled representation of the hidden states over time steps. Although this method is general enough and has achieved state-of-the-art results on many classification datasets, the concatenation representation is considered lack of representativeness.

Most recently, there are a couple of methods adopting a multi-layer Transformer \cite{vaswani2017attention}, which is a multi-headed self-attention based model, for the language model pre-training. OpenAI GPT (Generative Pre-trained Transformer) \cite{radford2018improving} uses a left-to-right Transformer and introduces the supervised fine-tuning with only one extra output layer. In this work, the authors propose a novel task-specific input transformations, which converts structured inputs into an ordered sequence that can fit into the pre-trained language model. Moreover, in the supervised fine-tuning, they also include language modeling as an auxiliary objective in order to improve generalization of the supervised model. Besides, the BERT (Bidirectional Encoder Representations from Transformers) model \cite{devlin2018bert} employs a bidirectional Transformer and is pre-trained using two novel unsupervised prediction tasks, namely Masked LM (MLM) and Next Sentence Prediction. MLM trains the LM with masking 15\% of tokens in each sequence at random, and Next Sentence Prediction pre-trains a binarized next sentence prediction task to make the model understand sentence relationships.

In this paper, we use ULMFiT as the main method for comparison for the following reasons. On one hand, although OpenAI GPT and BERT might perform better on different types of language understanding tasks, we only focus on the text classification tasks in this work as ULMFiT does. On the other hand, our goal is to demonstrate the effectiveness of the attention mechanism in fine-tuning. Although the transformer networks have strong ability to capture longer-range linguistic structure, we aim to use variants of long short-term memory (LSTM) network \cite{hochreiter1997long} in our LM due to its simplicity for implementation and efficiency in training.

\section{Framework}\label{sec:model}
\begin{figure}
\includegraphics[width=\textwidth]{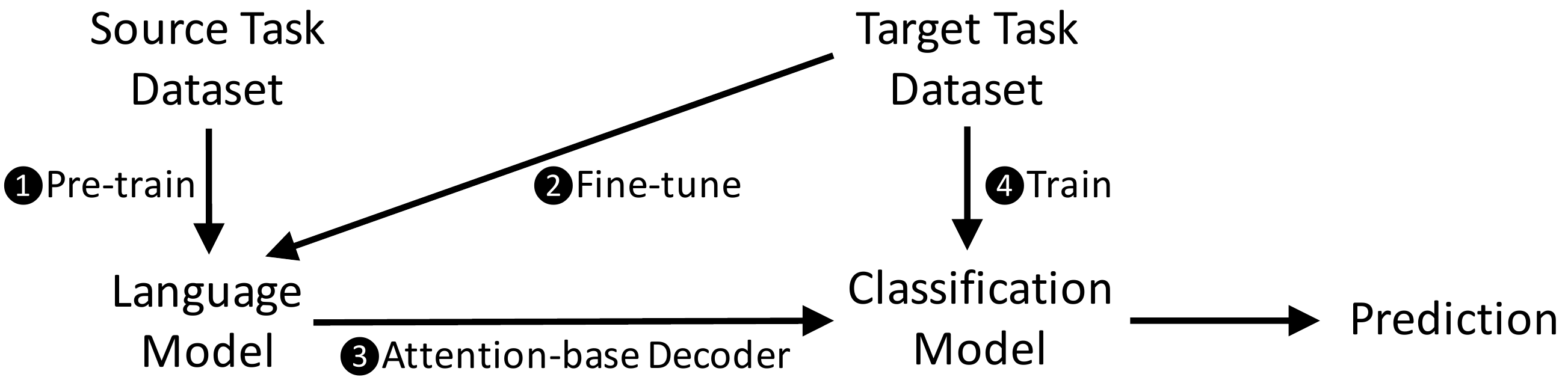}
\caption{High-level illustration of the proposed algorithm.}\label{fig:model}
\end{figure}

In this section, we will illustrate our training procedure, which is also shown in Figure \ref{fig:model}. Our training consists of four steps. We first pre-train a language model on a large scale text corpus. Then the pre-trained model is fine-tuned by the downstream classification dataset on unsupervised language modeling tasks. Moreover, we use an attention-based decoder to build our classification model and train the classifier using target task datasets.

\subsection{Unsupervised Pre-training}
Given a sequence of $N$ tokens $\mathcal{X}=(x_1, x_2, \ldots, x_N)$, a language model computes the probability of the sequence by modeling the probability of token $x_t$ given the history $(x_1, x_2, \ldots, x_{t-1})$:
\begin{equation}
p(x_1, x_2, \ldots, x_N)=\prod_{t=1}^N p(x_t | x_1, x_2, \ldots, x_{t-1})\,.
\end{equation}
Assume we have a vector $h_{t-1}\in\mathbb{R}^d$ encoding the history $(x_1, x_2, \ldots, x_t)$. Then the conditional probability of a token $x_t$ can be parametrized as
\begin{equation}\label{eq:prob}
p(x_t | x_1, x_2, \ldots, x_{t-1}) \propto \exp(U h_{t-1})\,,
\end{equation}
where $U$ is the weight matrix to be learned. In a recurrent neural network, the hidden state $h_t$ is usually computed from $h_{t-1}$ and $x_t$, namely
\beq
h_t = \Phi(x_t, h_{t-1})\,.
\eeq
In this paper, we aim to use AWD-LSTM \citep{merity2017regularizing} to model the conditional probability. The update of the hidden state $h_t$ from an LSTM can be defined as
\begin{equation}
\begin{aligned}
i_t &= \sigma (W_i x_t + U_i h_{t-1})\,, \\
f_t &= \sigma (W_f x_t + U_f h_{t-1})\,, \\
o_t &= \sigma (W_o x_t + U_o h_{t-1})\,, \\
\tilde{c}_t &= \tanh (W_c x_t + U_c h_{t-1})\,, \\
c_t &= i_t \odot \tilde{c}_t + f_t \odot \tilde{c}_{t-1}\,, \\
h_t &= o_t \odot \tanh(c_t)\,,
\end{aligned}
\end{equation} 
where $[W_i, W_f, W_o, W_c, U_i, U_f, U_o, U_c]$ are weight matrices, $h_t$ is the current exposed hidden state, $c_t$ is the current memory cell state, and $\odot$ denotes the element-wise multiplication. In order to prevent overfitting within the recurrent connections of an RNN, the AWD-LSTM performs the DropConnect \citep{wan2013regularization} on the hidden-to-hidden matrices $[U_i, U_f, U_o, U_c]$ within the LSTM.

For the training, we follow a standard language modeling objective to maximize the log-likelihood:
\beq\label{eq:l1}
L_1(\mathcal{X}) = \sum_{t=1}^N \log p(x_t | x_1, x_2, \ldots, x_{t-1}; \Theta)\,,
\eeq
where $p(\cdot)$ is defined in \eqref{eq:prob} and $\Theta$ is the parameter set from AWD-LSTM to be learned.

\subsection{End-to-end Fine-tuning}
Note that the data of the downstream classification task usually comes from a distribution that is different from the pre-training data. In order to apply the pre-trained language model, we follow \citep{howard2018universal} by introducing the target task LM fine-tuning using the target classification dataset without labels (Step 2 in Figure \ref{fig:model}). 

For classification, we adopt an attention-based encoder-decoder structure (Figure \ref{fig:model}, Step 3). Self-attention has been widely used in a variety of tasks including reading comprehension \cite{cheng2016long}, textual entailment \citep{parikh2016decomposable}, and abstractive summarization \citep{paulus2017deep}. As the encoder, our pre-trained model learns the contextualized features from inputs of the dataset. Then the hidden states over time steps, denoted as $H=\{h_1, h_2, \ldots, h_T\}$, can be viewed as the representation of the data to be classified, which are also the input of attention layer. Since we do not have any additional information from the decoder, we use the self-attention to extract the relevant aspects from the input states. Specifically, the alignment is computed as
\beq
u_t = \tanh(W_u h_t + b_u)\,,
\eeq
for $t=1,2,\ldots,T$, where $W_u$ and $b_u$ are the weight matrix and bias term to be learned. Then the alignment scores are given by the following soft-max function:
\beq
\alpha_t = \frac{\exp(W_a u_t)}{\sum_{i=1}^T \exp(W_a u_i)}\,.
\eeq
The final context vector, which is also the input of the decoder (classifier), is computed by
\beq
c = \sum_{t=1}^T \alpha_t u_t\,.
\eeq 

For the classifier, we follow \citep{howard2018universal} and the standard practice for CV classifiers, namely two additional linear blocks with batch normalization \cite{ioffe2015batch} and dropout \cite{srivastava2014dropout}, and ReLU activations for the intermediate layer and a soft-max activation for the output layer that calculates a probability distribution over target classes. Assume the output of the last linear block is $s_o$. Moreover, denote by $\mathcal{C}=\{c_1, c_2, \ldots, c_M\}=\mathcal{X}\times\mathcal{Y}$ the target classification data, where $c_i = (x^i, y^i)$, $x^i$ is the input sequence of tokens and $y^i$ is the corresponding label. Then the classification loss we use to train the model (Figure \ref{fig:model}, Step 4) can be computed by
\beq\label{eq:l2}
L_2(\mathcal{C}) = \sum_{(x,y)\in\mathcal{C}} \log p(y | x)\,,
\eeq
where
\beq
p(y | x) = p(y | x_1, x_2, \ldots, x_m) := \text{softmax}(W s_o)\,.
\eeq

However, for some large target datasets, such as Yelp and Sogou News, the LM fine-tuning (Figure \ref{fig:model}, Step 2) can take up to few days. Therefore, we can fine-tune the pre-trained model directly on the classification task, but in a sense of multi-task learning. More specifically, we combine the LM fine-tuning and classification fine-tuning in one objective optimization, which leads to an end-to-end fine-tuning. In other words, instead of Eq. \eqref{eq:l2}, we aim to maximize the following objective:
\beq\label{eq:total}
L(\mathcal{C}) = L_2(\mathcal{C}) + \lambda * L_1(\mathcal{X})\,,
\eeq
where $\lambda$ is the pre-defined weight. More discussion on the multi-task learning can be found in Section \ref{sec:multi-task}.

\section{Experiments}\label{sec:exp}
\subsection{Experimental Setup}
\begin{table}[t]
\caption{Statistics of datasets.}
\label{table:data}
\begin{center}
\begin{tabular}{ccccc}
\toprule
Dataset & \# classes & \# examples & Avg. sequence length & \# test  \\ \hline
AG & 4& 120k & 51 & 7.6k\\ 
DBpedia & 14 & 560k &61& 70k\\ 
Yelp-bi & 2 & 560k &163& 38k \\ 
Yelp-full & 5 &650k &164& 50k\\ 
IMDb & 2 & 25k &278 & 25k\\ 
Sogou & 5 & 450k &600 & 60k\\ 
\Xhline{1pt}
\end{tabular}
\end{center}
\vspace{.2em}
\end{table}
We evaluate our model on six widely studied datasets, with varying document length and number of classes. The statistics for each dataset are presented in Table \ref{table:data}. Note that the average sequence length is the average number of tokens after data pre-processing.

For topic classification, we evaluate on the large-scale Sogou news, AG news and DBpedia ontology datasets \cite{zhang2015character}. For sentiment analysis, we evaluate our approach on the binary movie review IMDb dataset \cite{maas2011learning} and the binary and five-class version of the Yelp review dataset \cite{zhang2015character}. In addition, we use the same pre-processing as in earlier work \cite{howard2018universal}.

We first pre-train our LM with different model architectures and datasets in order to demonstrate what factor of pre-training is significant to the downstream tasks performance. Then we demonstrate the effectiveness of our self-attention mechanism by comparing the classification performance with ULMFiT using the same pre-trained LM provided by ULMFiT in \cite{howard2018universal}. We should remark here that we can obtain similar accuracy with the LM pre-trained by our own. At last, we present the classification results from multi-task learning and discuss the trade-offs between classification accuracy and training efficiency.

\subsection{Unsupervised Pre-training}
\begin{table}[t]
\caption{Language Model pre-training and application to AG news classification.}
\label{table:pretrain}
\begin{center}
\begin{tabular}{ccccc}
\toprule
\multirow{2}{*}{Model} & Dataset  &  \multirow{2}{*}{\# examples} & Perplexity  &Accuracy   \\ 
&(pre-trained) &&(pre-trained) & (classification) \\\hline
LSTMP & One Bn Word& 1.1B & ~44 & 91.39\%\\ 
AWD-LSTM & WikiText-103 & 103.2M & ~69 & 92.83\%\\ 
AWD-LSTM & WikiText-2 & 2.1M & ~65 & 93.38\%  \\ 
\Xhline{1pt}
\end{tabular}
\end{center}
\end{table}

Regarding the language model pre-training, we have tried three different ways: 1) LSTM with projection layer (LSTMP) \cite{sak2014long} on One Billion Word Benchmark dataset; 2) AWD-LSTM on WikiText-103 dataset \cite{merity2016pointer}; and 3) AWD-LSTM on WikiText-2 dataset \cite{merity2016pointer}.

For LSTMP, we use word embedding dimension of 512, a one-layer LSTM with hidden size 2048 and projected to 512-dimensional output, while for AWD-LSTM we use word embedding dimension of 400 and a 3-layer AWD-LSTM with hidden size 1150. In order to evaluate the performance of pre-trained language models, we add an attention-based classifier on the top of LMs, and fine-tuned the model over AG's News dataset. The results of pre-training and classification fine-tuning are presented in Table \ref{table:pretrain}.

From the results in Table \ref{table:pretrain}, we can have the following observations. On one hand, although LSTMP on One Billion Word dataset performs best in the pre-training, it obtains the lowest accuracy in the classification task, which may indicate the problem of overfitting. In comparison to the size of target data (AG news), our language models are large enough (have enough parameters). In this case, AWD-LSTM is more suitable than LSTMP because it adopts DropConnect for weight matrices of hidden-to-hidden states, which mitigates the overfitting. 

On the other hand, pre-training on larger source datasets does not always improve downstream task performance. WikiText-2 is a subset of WikiText-103 and it is much smaller than WikiText-103 or One Billion Word dataset. But pre-training on WikiText-2 leads to the best performance among the three. We can see that the size of source data is not significant once we have a large enough pre-trained dataset. 
This observation indicates the possibility that when the source dataset is large enough, the performance of language modeling is a significant factor on transfer learning.

\subsection{Self-attention Mechanism}
We then carry out experiments to demonstrate that our self-attention mechanism is universally effective in the fine-tuning. In order to compare with ULMFiT, we use the pre-trained language model provided by ULMFiT and follow \cite{howard2018universal} to fine-tune the LM on target datasets before incorporating the classifier. Table \ref{table:results} shows the classification performance on various datasets.

\begin{table}[t]
\caption{Test error rates (\%) on different text classification datasets.}
\label{table:results}
\begin{center}
\begin{tabular}{ccccccc}
\toprule
\multirow{2}{*}{Dataset} & Error rate & Error rate & Error rate & Error rate & Improvement  & Improvement\\ 
& (from scratch)&(SOTA) & (ULMFiT) & (self-attention)& over ULMFiT &  over SOTA \\ \hline
AG & 6.53&5.29 (\cite{howard2018universal})& 5.54 & \textbf{5.17} & 6.68\% & 2.27\%\\ 
IMDb &9.86 &5.00 (\cite{howard2018universal}) & 5.08 & \textbf{4.59}& 9.65\% & 8.20\%\\ 
DBpedia &1.00 &0.84 (\cite{johnson2016convolutional}) & 0.87 &\textbf{0.80} & 8.05\% & 4.76\%\\ 
Yelp-bi & 2.91&2.64 (\cite{johnson2017deep}) & 2.37 &\textbf{1.97}& 16.88\% & 25.38\%   \\ 
Yelp-full & 31.11 &30.58 (\cite{johnson2017deep}) &30.73 &\textbf{28.86}& 6.05\% &5.59\%\\ 
Sogou & 2.50&1.84 (\cite{johnson2017deep}) & 2.26 &\textbf{1.69} & 25.22\% & 8.15\%\\ 
\Xhline{1pt}
\end{tabular}
\end{center}
\end{table}

In Table \ref{table:results}, the second column presents the results of training from scratch. The third column presents the previous state-of-the-art (SOTA) results that ever found in the literature. Note that ULMFiT \cite{howard2018universal} achieves all the SOTA results except on Sogou news with a combination of both forward and backward path LMs. However, in our experiments, the comparison is based on the single model. The error rates in the fourth column are the results we obtain by running the implementation from \citeauthor{howard2018universal}\footnote{https://github.com/fastai/fastai}. The last two columns computes the relative improvements of our methods over ULMFiT and previous SOTA results.

\begin{figure}
\vspace{.2em}
\centering
\subfigure[Business]{
        \includegraphics[width=1.1\linewidth]{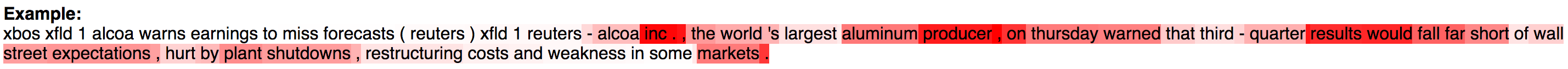}}
\subfigure[Science \& technology]{
        \includegraphics[width=1.1\linewidth]{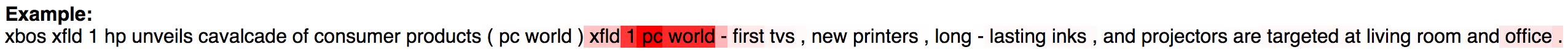}}
\subfigure[Sports]{
        \includegraphics[width=1.1\linewidth]{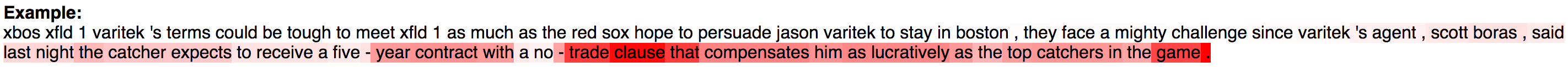}}        		
\subfigure[World]{
        \includegraphics[width=1.1\linewidth]{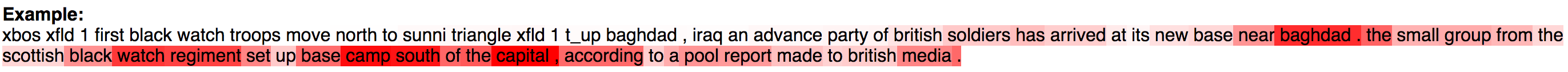}}        		
\caption{Examples of attention visualization on AG news with respect to different classes.}
\label{fig:vis}
\end{figure}

From Table \ref{table:results}, we can observe that by adding a self-attention layer, our model advances SOTA accuracies, as well as the performance of ULMFiT, on all the six datasets. The relative improvement can be as high as more than 25\%. In order to see the effectiveness of the attention layer more clearly, we visualize the attention scores with respect to the input texts on AG news. The randomly chosen examples of visualization with respect to different classes are given in Figure \ref{fig:vis}, where darker color means higher attention scores. Note that some tokens, such as <xbos> and <xfld 1>, represent the sentence tags that we used for data pre-processing and are not specific words in the original documents.

In our experiments, we have also tried ensemble self-attentions and multi-head attentions, the improvement is similar. Therefore, we adopted the simplest self-attention layer in our model architecture. 

\subsection{Multi-task Learning}\label{sec:multi-task}
As mentioned in Section 3.2, a separate target task LM fine-tuning is time costly. We list the LM fine-tuning time and classification fine-tuning time in Table \ref{table:multi} according to the experiments we presented in last section. From the table, we can see that the LM fine-tuning can take as much as more than half of the total training time.

\begin{figure}
\includegraphics[width=.95\textwidth]{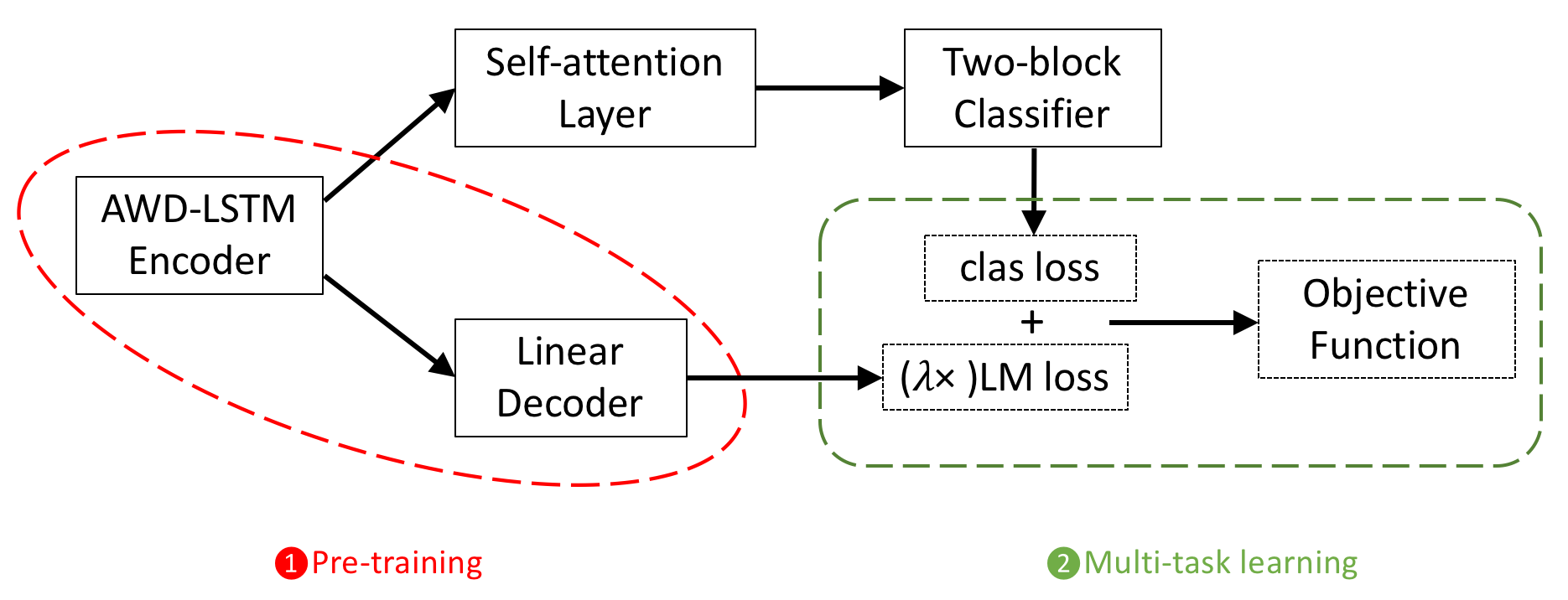}
\caption{Graphic illustration of the multi-task learning algorithm.}\label{fig:model-multi}
\end{figure}

In order to make the training more efficient, we can fine-tune the two parts together, which leads to a multi-task learning with the objective Eq. \eqref{eq:total}. Figure \ref{fig:model-multi} showcases the structure of our multi-task learning algorithm, which consists of two steps. We first pre-train a language model on a large scale text corpus. We then use two decoders to build the target model, one attention-base decoder for the classifier and one simple linear-block decoder for incorporating the language modeling loss. In the experiments, we choose the weight $\lambda$ in Eq. \eqref{eq:total} to be $0.1$.

We test the multi-task learning on AG news. The total training time until convergence is reduced from 6.5 hours to 3.5 hours. However, we get an error rate of 5.49\%, which is slightly better than ULMFiT but not as good as SOTA result. Therefore, customers should be aware that by using the multi-task learning framework, we have to trade off some classification accuracy for the training efficiency.

\begin{table}[!t]
\caption{Fine-tuning time of different datasets.}
\label{table:multi}
\begin{center}
\begin{tabular}{cccc}
\toprule
\multirow{2}{*}{Dataset} & LM fine-tuning & classification fine-tuning & Ratio of LM  \\ 
& time (hours) & time (hours) & fine-tuning\\ \hline
AG & 1.5 & 5 & 23.08\% \\
IMDb & 5 & 4 & 55.56\% \\
DBpedia & 23 & 30 & 43.40\%\\
Yelp-bi & 52 & 55 & 48.60\% \\
Yelp-full & 56 & 87 & 39.16\% \\
Sogou & 72 & 209 & 25.62\% \\
\Xhline{1pt}
\end{tabular}
\end{center}
\end{table}

\section{Conclusion}\label{sec:conclusion}
In this work, we have proposed an attention-based fine-tuning algorithm which provides a reliable and easy-to-use feature extractor from the pre-trained language model and uses those features for downstream text classification tasks. The performance of the proposed algorithm advances state-of-the-art methods on various benchmark datasets. With this algorithm, the customers can use the given language model and fine-tune the target model by their own data. In addition, the customers can also adopt another version of our algorithm, \textit{i.e.}, the approach of multi-task learning, for faster training if they allow a slight reduction of the model performance.

\small
\bibliography{lmas}

\begin{thebibliography}{32}
\providecommand{\natexlab}[1]{#1}
\providecommand{\url}[1]{\texttt{#1}}
\expandafter\ifx\csname urlstyle\endcsname\relax
  \providecommand{\doi}[1]{doi: #1}\else
  \providecommand{\doi}{doi: \begingroup \urlstyle{rm}\Url}\fi

\bibitem[Arisoy et~al.(2012)Arisoy, Sainath, Kingsbury, and
  Ramabhadran]{arisoy2012deep}
Arisoy, Ebru, Sainath, Tara~N, Kingsbury, Brian, and Ramabhadran, Bhuvana.
\newblock Deep neural network language models.
\newblock In \emph{Proceedings of the NAACL-HLT 2012 Workshop: Will We Ever
  Really Replace the N-gram Model? On the Future of Language Modeling for HLT},
  pp.\  20--28. Association for Computational Linguistics, 2012.

\bibitem[Chelba et~al.(2013)Chelba, Mikolov, Schuster, Ge, Brants, Koehn, and
  Robinson]{chelba2013one}
Chelba, Ciprian, Mikolov, Tomas, Schuster, Mike, Ge, Qi, Brants, Thorsten,
  Koehn, Phillipp, and Robinson, Tony.
\newblock One billion word benchmark for measuring progress in statistical
  language modeling.
\newblock \emph{arXiv preprint arXiv:1312.3005}, 2013.

\bibitem[Cheng et~al.(2016)Cheng, Dong, and Lapata]{cheng2016long}
Cheng, Jianpeng, Dong, Li, and Lapata, Mirella.
\newblock Long short-term memory-networks for machine reading.
\newblock \emph{arXiv preprint arXiv:1601.06733}, 2016.

\bibitem[Devlin et~al.(2018)Devlin, Chang, Lee, and Toutanova]{devlin2018bert}
Devlin, Jacob, Chang, Ming-Wei, Lee, Kenton, and Toutanova, Kristina.
\newblock Bert: Pre-training of deep bidirectional transformers for language
  understanding.
\newblock \emph{arXiv preprint arXiv:1810.04805}, 2018.

\bibitem[Filippova et~al.(2015)Filippova, Alfonseca, Colmenares, Kaiser, and
  Vinyals]{filippova2015sentence}
Filippova, Katja, Alfonseca, Enrique, Colmenares, Carlos~A, Kaiser, Lukasz, and
  Vinyals, Oriol.
\newblock Sentence compression by deletion with lstms.
\newblock In \emph{Proceedings of the 2015 Conference on Empirical Methods in
  Natural Language Processing}, pp.\  360--368, 2015.

\bibitem[Hochreiter \& Schmidhuber(1997)Hochreiter and
  Schmidhuber]{hochreiter1997long}
Hochreiter, Sepp and Schmidhuber, J{\"u}rgen.
\newblock Long short-term memory.
\newblock \emph{Neural computation}, 9\penalty0 (8):\penalty0 1735--1780, 1997.

\bibitem[Howard \& Ruder(2018)Howard and Ruder]{howard2018universal}
Howard, Jeremy and Ruder, Sebastian.
\newblock Universal language model fine-tuning for text classification.
\newblock In \emph{Proceedings of the 56th Annual Meeting of the Association
  for Computational Linguistics (Volume 1: Long Papers)}, volume~1, pp.\
  328--339, 2018.

\bibitem[Ioffe \& Szegedy(2015)Ioffe and Szegedy]{ioffe2015batch}
Ioffe, Sergey and Szegedy, Christian.
\newblock Batch normalization: Accelerating deep network training by reducing
  internal covariate shift.
\newblock \emph{arXiv preprint arXiv:1502.03167}, 2015.

\bibitem[Johnson \& Zhang(2016)Johnson and Zhang]{johnson2016convolutional}
Johnson, Rie and Zhang, Tong.
\newblock Convolutional neural networks for text categorization: Shallow
  word-level vs. deep character-level.
\newblock \emph{arXiv preprint arXiv:1609.00718}, 2016.

\bibitem[Johnson \& Zhang(2017)Johnson and Zhang]{johnson2017deep}
Johnson, Rie and Zhang, Tong.
\newblock Deep pyramid convolutional neural networks for text categorization.
\newblock In \emph{Proceedings of the 55th Annual Meeting of the Association
  for Computational Linguistics (Volume 1: Long Papers)}, volume~1, pp.\
  562--570, 2017.

\bibitem[Maas et~al.(2011)Maas, Daly, Pham, Huang, Ng, and
  Potts]{maas2011learning}
Maas, Andrew~L, Daly, Raymond~E, Pham, Peter~T, Huang, Dan, Ng, Andrew~Y, and
  Potts, Christopher.
\newblock Learning word vectors for sentiment analysis.
\newblock In \emph{Proceedings of the 49th annual meeting of the association
  for computational linguistics: Human language technologies-volume 1}, pp.\
  142--150. Association for Computational Linguistics, 2011.

\bibitem[McCann et~al.(2017)McCann, Bradbury, Xiong, and
  Socher]{mccann2017learned}
McCann, Bryan, Bradbury, James, Xiong, Caiming, and Socher, Richard.
\newblock Learned in translation: Contextualized word vectors.
\newblock In \emph{Advances in Neural Information Processing Systems}, pp.\
  6294--6305, 2017.

\bibitem[Merity et~al.(2016)Merity, Xiong, Bradbury, and
  Socher]{merity2016pointer}
Merity, Stephen, Xiong, Caiming, Bradbury, James, and Socher, Richard.
\newblock Pointer sentinel mixture models.
\newblock \emph{arXiv preprint arXiv:1609.07843}, 2016.

\bibitem[Merity et~al.(2017)Merity, Keskar, and Socher]{merity2017regularizing}
Merity, Stephen, Keskar, Nitish~Shirish, and Socher, Richard.
\newblock Regularizing and optimizing lstm language models.
\newblock \emph{arXiv preprint arXiv:1708.02182}, 2017.

\bibitem[Mikolov et~al.(2010)Mikolov, Karafi{\'a}t, Burget, {\v{C}}ernock{\`y},
  and Khudanpur]{mikolov2010recurrent}
Mikolov, Tom{\'a}{\v{s}}, Karafi{\'a}t, Martin, Burget, Luk{\'a}{\v{s}},
  {\v{C}}ernock{\`y}, Jan, and Khudanpur, Sanjeev.
\newblock Recurrent neural network based language model.
\newblock In \emph{Eleventh Annual Conference of the International Speech
  Communication Association}, 2010.

\bibitem[Parikh et~al.(2016)Parikh, T{\"a}ckstr{\"o}m, Das, and
  Uszkoreit]{parikh2016decomposable}
Parikh, Ankur~P, T{\"a}ckstr{\"o}m, Oscar, Das, Dipanjan, and Uszkoreit, Jakob.
\newblock A decomposable attention model for natural language inference.
\newblock \emph{arXiv preprint arXiv:1606.01933}, 2016.

\bibitem[Paulus et~al.(2017)Paulus, Xiong, and Socher]{paulus2017deep}
Paulus, Romain, Xiong, Caiming, and Socher, Richard.
\newblock A deep reinforced model for abstractive summarization.
\newblock \emph{arXiv preprint arXiv:1705.04304}, 2017.

\bibitem[Pennington et~al.(2014)Pennington, Socher, and
  Manning]{pennington2014glove}
Pennington, Jeffrey, Socher, Richard, and Manning, Christopher.
\newblock Glove: Global vectors for word representation.
\newblock In \emph{Proceedings of the 2014 conference on empirical methods in
  natural language processing (EMNLP)}, pp.\  1532--1543, 2014.

\bibitem[Peters et~al.(2017)Peters, Ammar, Bhagavatula, and
  Power]{peters2017semi}
Peters, Matthew~E, Ammar, Waleed, Bhagavatula, Chandra, and Power, Russell.
\newblock Semi-supervised sequence tagging with bidirectional language models.
\newblock \emph{arXiv preprint arXiv:1705.00108}, 2017.

\bibitem[Peters et~al.(2018)Peters, Neumann, Iyyer, Gardner, Clark, Lee, and
  Zettlemoyer]{peters2018deep}
Peters, Matthew~E, Neumann, Mark, Iyyer, Mohit, Gardner, Matt, Clark,
  Christopher, Lee, Kenton, and Zettlemoyer, Luke.
\newblock Deep contextualized word representations.
\newblock \emph{arXiv preprint arXiv:1802.05365}, 2018.

\bibitem[Radford et~al.(2017)Radford, Jozefowicz, and
  Sutskever]{radford2017learning}
Radford, Alec, Jozefowicz, Rafal, and Sutskever, Ilya.
\newblock Learning to generate reviews and discovering sentiment.
\newblock \emph{arXiv preprint arXiv:1704.01444}, 2017.

\bibitem[Radford et~al.(2018)Radford, Narasimhan, Salimans, and
  Sutskever]{radford2018improving}
Radford, Alec, Narasimhan, Karthik, Salimans, Tim, and Sutskever, Ilya.
\newblock Improving language understanding by generative pre-training.
\newblock 2018.

\bibitem[Rush et~al.(2015)Rush, Chopra, and Weston]{rush2015neural}
Rush, Alexander~M, Chopra, Sumit, and Weston, Jason.
\newblock A neural attention model for abstractive sentence summarization.
\newblock \emph{arXiv preprint arXiv:1509.00685}, 2015.

\bibitem[Sak et~al.(2014)Sak, Senior, and Beaufays]{sak2014long}
Sak, Ha{\c{s}}im, Senior, Andrew, and Beaufays, Fran{\c{c}}oise.
\newblock Long short-term memory based recurrent neural network architectures
  for large vocabulary speech recognition.
\newblock \emph{arXiv preprint arXiv:1402.1128}, 2014.

\bibitem[Schwenk et~al.(2012)Schwenk, Rousseau, and Attik]{schwenk2012large}
Schwenk, Holger, Rousseau, Anthony, and Attik, Mohammed.
\newblock Large, pruned or continuous space language models on a gpu for
  statistical machine translation.
\newblock In \emph{Proceedings of the NAACL-HLT 2012 Workshop: Will We Ever
  Really Replace the N-gram Model? On the Future of Language Modeling for HLT},
  pp.\  11--19. Association for Computational Linguistics, 2012.

\bibitem[Shen et~al.(2017)Shen, Zhou, Long, Jiang, Pan, and
  Zhang]{shen2017disan}
Shen, Tao, Zhou, Tianyi, Long, Guodong, Jiang, Jing, Pan, Shirui, and Zhang,
  Chengqi.
\newblock Disan: Directional self-attention network for rnn/cnn-free language
  understanding.
\newblock \emph{arXiv preprint arXiv:1709.04696}, 2017.

\bibitem[Srivastava et~al.(2014)Srivastava, Hinton, Krizhevsky, Sutskever, and
  Salakhutdinov]{srivastava2014dropout}
Srivastava, Nitish, Hinton, Geoffrey, Krizhevsky, Alex, Sutskever, Ilya, and
  Salakhutdinov, Ruslan.
\newblock Dropout: a simple way to prevent neural networks from overfitting.
\newblock \emph{The Journal of Machine Learning Research}, 15\penalty0
  (1):\penalty0 1929--1958, 2014.

\bibitem[Vaswani et~al.(2013)Vaswani, Zhao, Fossum, and
  Chiang]{vaswani2013decoding}
Vaswani, Ashish, Zhao, Yinggong, Fossum, Victoria, and Chiang, David.
\newblock Decoding with large-scale neural language models improves
  translation.
\newblock In \emph{Proceedings of the 2013 Conference on Empirical Methods in
  Natural Language Processing}, pp.\  1387--1392, 2013.

\bibitem[Vaswani et~al.(2017)Vaswani, Shazeer, Parmar, Uszkoreit, Jones, Gomez,
  Kaiser, and Polosukhin]{vaswani2017attention}
Vaswani, Ashish, Shazeer, Noam, Parmar, Niki, Uszkoreit, Jakob, Jones, Llion,
  Gomez, Aidan~N, Kaiser, {\L}ukasz, and Polosukhin, Illia.
\newblock Attention is all you need.
\newblock In \emph{Advances in Neural Information Processing Systems}, pp.\
  5998--6008, 2017.

\bibitem[Wan et~al.(2013)Wan, Zeiler, Zhang, Le~Cun, and
  Fergus]{wan2013regularization}
Wan, Li, Zeiler, Matthew, Zhang, Sixin, Le~Cun, Yann, and Fergus, Rob.
\newblock Regularization of neural networks using dropconnect.
\newblock In \emph{International Conference on Machine Learning}, pp.\
  1058--1066, 2013.

\bibitem[Zhang et~al.(2015)Zhang, Zhao, and LeCun]{zhang2015character}
Zhang, Xiang, Zhao, Junbo, and LeCun, Yann.
\newblock Character-level convolutional networks for text classification.
\newblock In \emph{Advances in neural information processing systems}, pp.\
  649--657, 2015.

\bibitem[Zhou et~al.(2017)Zhou, Yang, Wei, and Zhou]{zhou2017selective}
Zhou, Qingyu, Yang, Nan, Wei, Furu, and Zhou, Ming.
\newblock Selective encoding for abstractive sentence summarization.
\newblock \emph{arXiv preprint arXiv:1704.07073}, 2017.

\end{thebibliography}
\bibliographystyle{icml2018}

\end{document}